\documentclass[10pt,twocolumn,letterpaper]{article}
\usepackage{cvpr}              
\usepackage{multirow}
\usepackage{booktabs, multirow} 
\usepackage{soul}
\usepackage{colortbl} 
\usepackage{changepage,threeparttable} 
\usepackage[export]{adjustbox}
\usepackage{multirow}
\newcommand{\model}{ReVisionLLM}

\definecolor{cvprblue}{rgb}{0.21,0.49,0.74}
\usepackage[pagebackref,breaklinks,colorlinks,citecolor=cvprblue]{hyperref}
\usepackage{multirow}
\usepackage{multicol}
\usepackage{booktabs}
\usepackage{graphicx}
\usepackage[export]{adjustbox}
\usepackage{pifont}
\usepackage{amssymb}
\usepackage{amstext}
\usepackage{amsmath}
\usepackage[bb=boondox]{mathalfa}
\usepackage[dvipsnames]{xcolor}
\usepackage{float}
\usepackage{graphicx}
\usepackage{amsmath}
\usepackage{amssymb}
\usepackage{booktabs}
\usepackage{siunitx}
\usepackage{bm}
\usepackage{subcaption}
\usepackage{soul}
\usepackage{array}
\usepackage{pifont}
\usepackage[accsupp]{axessibility}

\newcolumntype{P}[1]{>{\centering\arraybackslash}p{#1}}

\definecolor{orange}{RGB}{255,127,0}
\title{\model: Recursive Vision-Language Model for \\Temporal Grounding in Hour-Long Videos}
\author{
Tanveer Hannan$^{1,2}\thanks{Corresponding author: hannan@dbs.ifi.lmu.de}$ \quad
Md Mohaiminul Islam$^{3}$ \quad
Jindong Gu$^{4}$ \quad
Thomas Seidl$^{1,2}$ \quad
Gedas Bertasius$^3$\\ 
$^1$ LMU Munich \quad
$^2$ MCML \quad
$^3$ UNC Chapel Hill \quad
$^4$ University of Oxford\\
}

\begin{document}
\maketitle
\begin{abstract}
Large language models (LLMs) excel at retrieving information from lengthy text, but their vision-language counterparts (VLMs) face difficulties with hour-long videos, especially for temporal grounding. Specifically, these VLMs are constrained by frame limitations, often losing essential temporal details needed for accurate event localization in extended video content. We propose \model, a recursive vision-language model designed to locate events in hour-long videos. Inspired by human search strategies, our model initially targets broad segments of interest, progressively revising its focus to pinpoint exact temporal boundaries. Our model can seamlessly handle videos of vastly different lengths—from minutes to hours. We also introduce a hierarchical training strategy that starts with short clips to capture distinct events and progressively extends to longer videos. To our knowledge, \model\ is the first VLM capable of temporal grounding in hour-long videos, outperforming previous state-of-the-art methods across multiple datasets by a significant margin (e.g., +2.6\% R1@0.1 on MAD). The code is available at \href{https://github.com/Tanveer81/ReVisionLLM}{https://github.com/Tanveer81/ReVisionLLM}
\vspace{-0.5cm}
\end{abstract}    
\section{Introduction}
\label{sec:intro}
Large language models (LLMs) are particularly adept at handling extensive text documents, such as full-length books, and retrieving relevant information~\cite{1GregKam40, 1GregKam32, gkamradt71, liu2023lost, ivgi2023efficient}. However, achieving similar capabilities in video, i.e., locating fine-grained temporal events in hour-long videos, remains a critical challenge. This task, known as long-video temporal grounding, requires accurately identifying the start and end of events based on a user's textual query. This capability could be paramount for video content search, sports analytics, surveillance, and many other applications. However, current vision-language models (VLMs) struggle with this demanding task. 

\begin{figure}
    \centering
    \includegraphics[width=\linewidth]{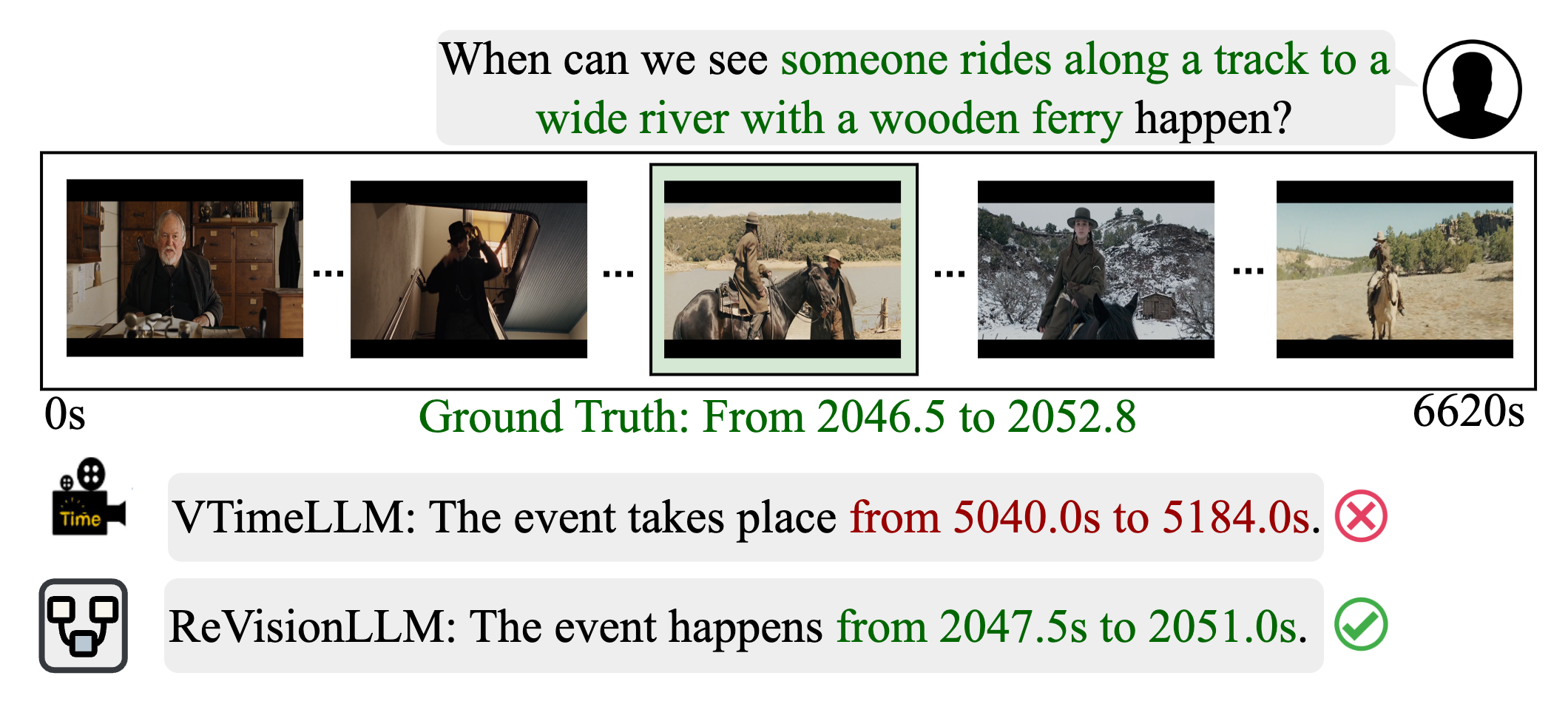}
    \vspace{-6mm}
    \caption{Existing vision-language models (VLMs) such as VTimeLLM~\cite{huang2024vtimellm} are not equipped to process hour-long videos effectively and struggle to pinpoint precise temporal boundaries for events within extended video durations. In contrast, \model\ is the first VLM designed to address this limitation, enabling accurate temporal grounding in hour-long video content.}
    \label{fig:intro}
    \vspace{-4mm}
\end{figure}

Recently, non-LLM-based models~\cite{pan2023scanning, hannan2023rgnet, hou2022cone, mu2024snag} have made progress in long temporal video grounding. However, these methods typically involve multiple networks and complex post-processing steps. Additionally, these models generally lack the flexibility to handle textual user instructions. In contrast, the recent VLMs~\cite{huang2024vtimellm, ren2024timechat, qian2024momentor, huang2024lita} can effectively process textual user queries but are ineffective for temporal localization in long videos  (Fig. \ref{fig:intro}). In particular, such VLM-based approaches tend to underperform even on short video (e.g., 2 minutes) localization tasks compared to non-LLM approaches~\cite{moon2023query, lin2023univtg, lei2021detecting, liu2022umt}.

Extending LLM-based solutions to hour-long video inputs for temporal localization presents several important challenges. Video data is much denser than text, leading to a massive number of input tokens for the LLM. To handle this, many VLMs downsample frames
and operate on a limited number of frames~\cite{lin2023video, huang2024vtimellm, jin2024chat, zhang2023video, xu2024pllava}, which leads to a significant loss of information in long videos. Moreover, training VLMs on hour-long videos requires immense memory and computational resources, which presents practical challenges for scalability and training efficiency. Furthermore, current VLMs often exhibit poor confidence calibration~\cite{kostumov2024uncertainty,oh2023towards}, leading to frequent false positives with high confidence. This issue is amplified in long videos, where distinguishing actual events from numerous false detections becomes increasingly difficult. 

\begin{figure}[!t]
    \centering
    \includegraphics[width=\linewidth]{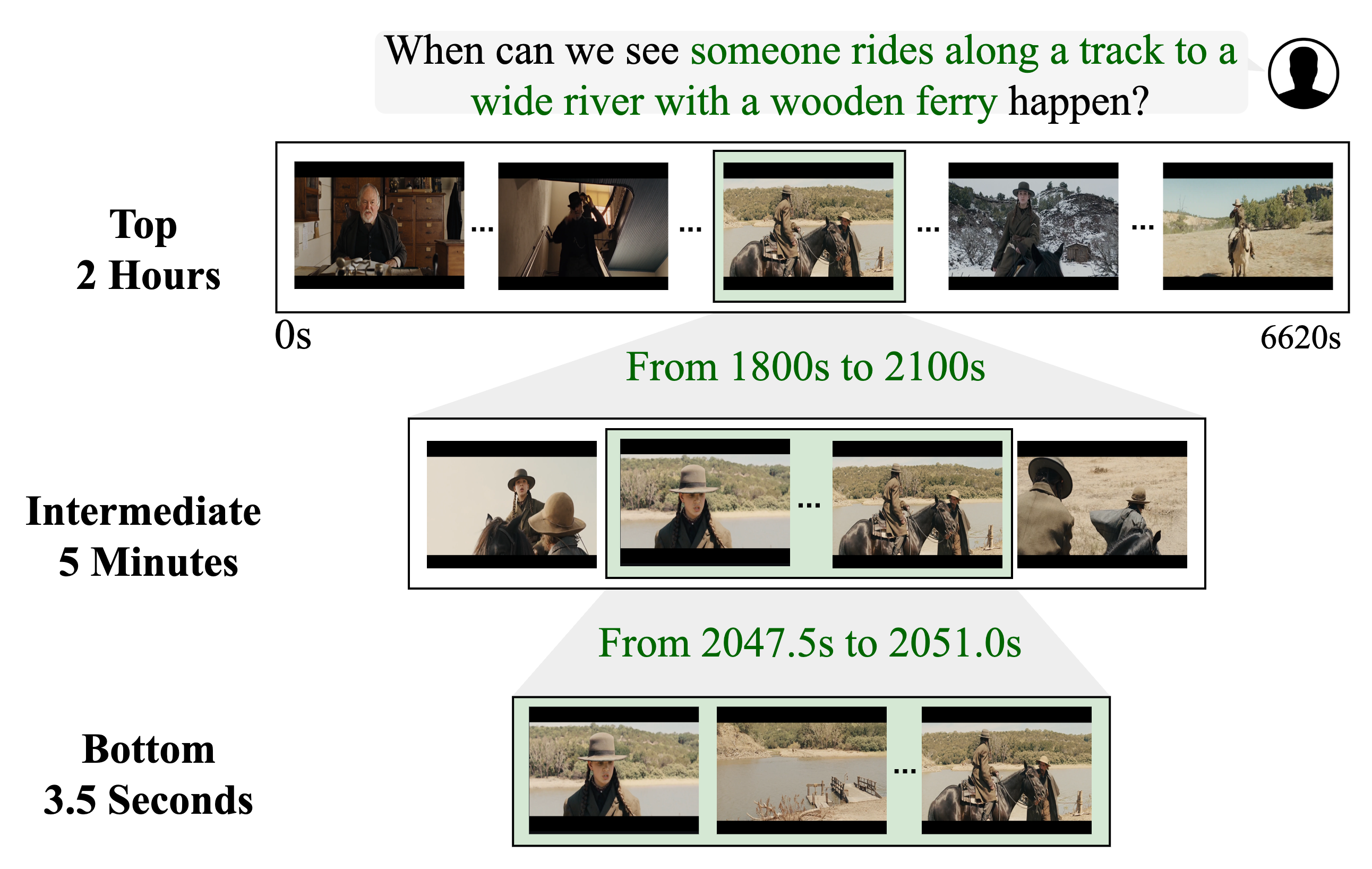}
    \vspace{-6mm}
    \caption{\textbf{Recursive Video Grounding.} \model\ is a recursive vision-language model designed for localizing events in hour-long videos. Inspired by human search strategies, it first scans the entire video to identify relevant intermediate segments and then zooms in to precisely locate event boundaries. Here, we show one intermediate hierarchy for brevity.}
    \label{fig:main}
    \vspace{-4mm}
\end{figure}

To address these challenges, we introduce the Recursive Vision-Language Model \model, a VLM with hierarchical perception that processes videos recursively. Cognitive studies~\cite{bourke2013functional, wolfe2017five} suggest that when searching content, humans maintain a mental representation of the target and direct attention to the most promising areas, refining their search. In line with these principles, given a long video input, our recursive model first identifies broad video segments of interest and then progressively revises its focus, narrowing in on the event's exact temporal boundaries.  In Fig.~\ref{fig:main}, we show the operating principle of our model. At the top hierarchy, the model operates broadly, identifying relevant segments (e.g., 5 minutes) from a 2-hour-long video. As it moves down the intermediate hierarchies, it narrows its focus to increasingly fine-grained temporal segments at the lowest hierarchy, pinpointing precise event boundaries (e.g., 3.5 seconds). Such a recursive processing structure of our model allows it to scale effectively to hour-long videos. 

We first train our model on short video segments, then progress to training on hour-long videos. In the short video training phase, we introduce contrastive segments (i.e., video clips that do not contain the queried event) to improve confidence calibration. This helps the model learn to identify both the presence and absence of events, enhancing its confidence in visual input and aiding in accurate event localization within long videos. For efficient training on hour-long videos, we employ a temporal feature reduction strategy that compresses video segments into compact representations. This method reduces the input tokens required by the LLM. As a result, our model achieves both high accuracy and efficiency, making it well-suited for analyzing lengthy videos. Our contributions can be summarized as follows: 

\begin{itemize}
    \item We extend the existing VLMs to enable temporal grounding capabilities in \textbf{hour-long videos}.
    \item We propose a vision-language model that recursively processes hour-long videos for effective and efficient hour-long video processing.
    \item We propose a progressive training strategy, where the model is first trained to identify events in short video segments, then progressively scales to hour-long videos, enabling it to effectively handle longer, more complex video sequences. 
    \item Our model significantly outperforms previous state-of-the-art approaches, surpassing specialized models and other Vision-Language Models (VLMs) on multiple datasets by a substantial margin. For instance, \model\ outperforms the previous best method ~\cite{hannan2023rgnet} by \textbf{2.6\%} R1@.1 on the MAD ~\cite{soldan2022mad} dataset. Moreover, our model can efficiently solve this task, processing, on average, 43\% fewer frames compared to the the existing VLM model~\cite{huang2024vtimellm}.
\end{itemize}

\section{Related Work}
\label{sec:related}

\vspace{1mm}
\noindent\textbf{Vision Language Models.} VLMs excel in tasks such as video summarization~\cite{zhao2023learning, zhang2024simplellmframeworklongrange}, decision-making~\cite{wang2023chatvideo, gao2023assistgpt}, captioning and general question-answering~\cite{chen2023videollm, song2024moviechat, zhang2023video}, temporal localization and object trajectory detection~\cite{huang2024vtimellm, ren2024timechat, huang2024lita, yu2024self, li2024groundinggpt, wang2024hawkeye}. Existing VLMs integrate visual input by either training-free methods~\cite{xu2024slowfast}, full fine-tuning~\cite{yang2023vid2seq}, or adapter fine-tuning~\cite{liu2023llava, li2023blip, hu2021lora, huang2024vtimellm, ren2024timechat}. However, most VLMs are limited to short videos and lack extensive temporal awareness. Our work extends VLMs to enable capabilities for temporal localization in hour-long videos. We contribute a novel adapter fine-tuning approach for hierarchical perception. Unlike prior models that aggregate frame features through spatial-temporal pooling~\cite{maaz2023video} or align visual-text embeddings~\cite{li2023videochat}, we introduce a temporal feature reduction method to effectively scale training to hour-long videos.

\vspace{1mm}
\noindent\textbf{Temporal Grounding VLMs.}
Most existing VLMs face challenges with temporal grounding. Recent models~\cite{huang2024vtimellm, ren2024timechat, huang2024lita, wang2024hawkeye, qian2024momentor, yu2024self, li2024groundinggpt, wang2024omnivid, tang2024avicuna, qu2024chatvtg} have been specifically developed to address this issue using specialized architectures and datasets. For example, VTimeLLM~\cite{huang2024vtimellm} proposes a temporal fine-tuning stage, TimeChat~\cite{ren2024timechat} integrates timestamps with visual features, LITA~\cite{huang2024lita} introduces time tokens for temporal understanding, and Hawkeye~\cite{wang2024hawkeye} uses dense video captions for segment matching. However, these models are limited by their training context length, which confines them to short video segments and hinders their ability to handle the complex temporal relationships and redundancies found in longer videos. Although they can process arbitrary user queries, they generally underperform compared to traditional, non-LLM-based models. We overcome these limitations by introducing a new recursive vision-language model that enables long video processing.

\vspace{1mm}
\noindent\textbf{Long Video Temporal Grounding.} Recent advancements in hour-long video temporal grounding have leveraged datasets like MAD~\cite{soldan2022mad} and VidChapters7M~\cite{yang2023vidchapters7mvideochaptersscale}. These methods typically follow a two-stage approach: proposal-free methods~\cite{zhang2020learning, zhang2020span, soldan2021vlg, barrios2023localizing, liu2022reler} segment videos to predict candidate moments and rank them, while proposal-based methods~\cite{pan2023scanning, hou2022cone} generate proposal clips or anchors for grounding models. Approaches like CONE~\cite{hou2022cone} and M-Guidance~\cite{barrios2023localizing} utilize detection transformers~\cite{carion2020end} to enhance grounding, while RGNet~\cite{hannan2023rgnet} unifies clip retrieval~\cite{msrvtt} and grounding with an end-to-end transformer-based approach. Recently, SnAG~\cite{mu2024snag} introduced late fusion techniques for more efficient processing. Prior models lack the instruction-following capability and rely on techniques that match video and text inputs. In contrast, our work integrates VLMs, which naturally allows our model to follow instructions and process textual user queries in the context of a long temporal grounding framework.

\section{Method}
\label{sec:method}
\begin{figure*}[!t]
    \centering
    \includegraphics[width=\linewidth]{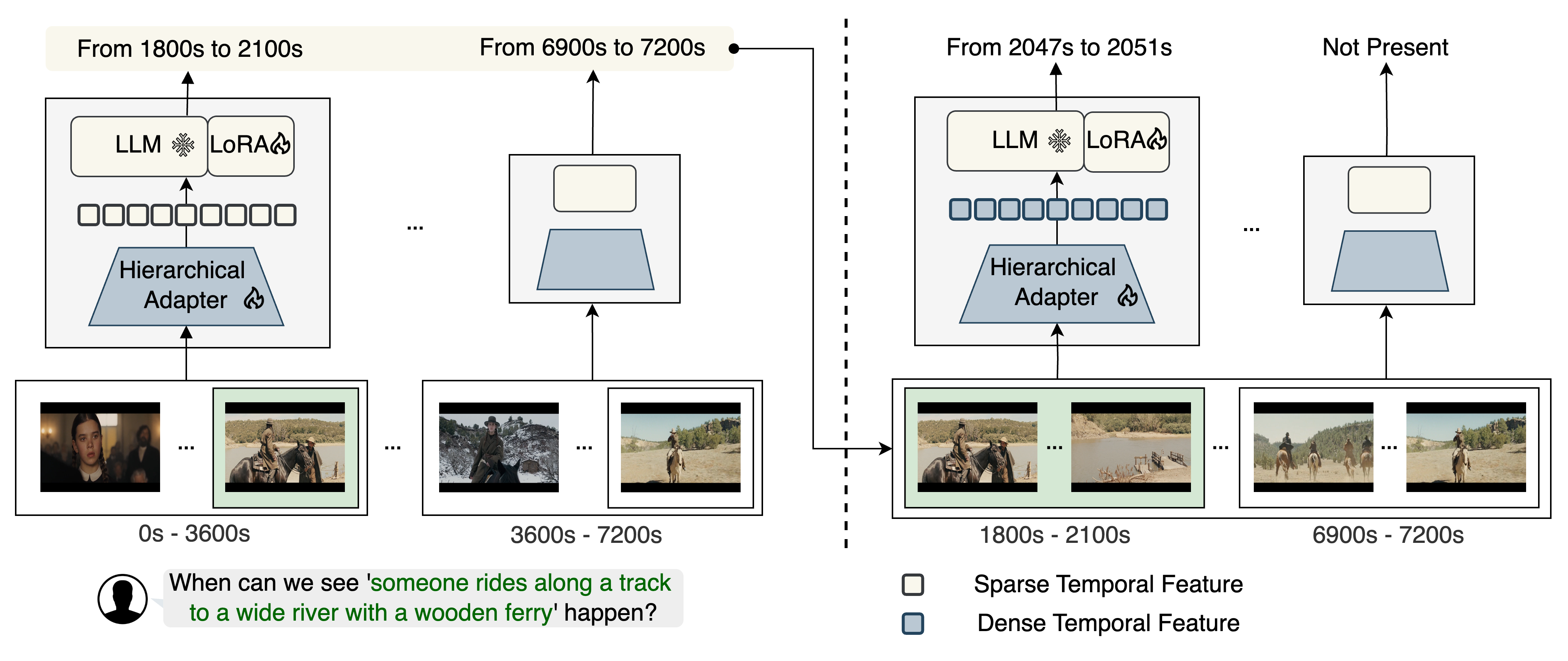}
    \vspace{-6mm}
    \caption{\textbf{The \model\ model.} \textbf{(Left)} First, we detect segments (e.g., a few minutes) from an hour-long video using sparse temporal features produced by the Hierarchical Adapter. \textbf{(Right)} Then \model\ produces a precise temporal boundary using dense temporal features within the predicted segments. Note that the green box represents the same event boundary in both sub-figures, zooming in from left to right. The multimodal encoder is omitted for simplicity.}
    \vspace{-4mm}
    \label{fig:method}
\end{figure*}

\subsection{Problem Overview}
\label{sec: task definition}

Given a long, untrimmed video input and an event defined by a text query, we aim to predict the precise temporal boundary of the event. Formally, as our inputs, we consider a long-range video sequence $V=[v^{t}]_{t=1,\hdots, T}$ comprised of $T$ RGB frames, where $v^{t}$ is the $t^{th}$ frame. The event is defined by a query sentence $S$ with $N_s$ words where the sentence corresponds to a target event's start ($s$) and end ($e$) times, denoted as \( {\tau = {(s,e)}} \). 

\subsection{The \model\ Model} 

We now describe our proposed \model\ model, which contains three high-level components: (1) a Multimodal Encoder, (2) a Hierarchical Adapter, and (3) a Large Language Model. We illustrate our approach in Fig. \ref{fig:method} and describe each component below.

\vspace{1mm}
\noindent\textbf{Multimodal Encoder.} We utilize an off-the-shelf video encoder (e.g., CLIP ViT-L/14~\cite{xue2022clip}) to extract features from an hour-long video. To reduce the input context length and capture the global properties of long video inputs, we extract global features (e.g., CLS token) for each frame, $f^{t} \in \mathbb{R}^D$

, where $D$ represents the feature dimension. These CLS tokens form a set of temporal features, ${F}=[f^{t}]_{t=1,\hdots, T}$ for the whole video. We use the same CLIP ViT-L/14 text encoder to extract textual features $Q \in \mathbb{R}^{N_s \times D}$ of the query sentence $S$. 

\vspace{1mm}
\noindent\textbf{Hierarchical Adapter.}  
While LLMs excel at retrieving information from long text documents, performing retrieval from a large number of video frames remains challenging, limiting their effectiveness for visual grounding. To address this, we employ a recursive approach that processes hour-long videos at different temporal resolutions. Initially, the entire video is processed as a whole to identify segments of interest (Fig. \ref{fig:main}-top). Next, these segments undergo finer analysis to pinpoint precise event boundaries (Fig. \ref{fig:main}-bottom). At the bottom level, we retain the original temporal resolution, while higher levels use compressed representations to maintain manageable visual input lengths for the LLM. Specifically, our Hierarchical Adapter projects initial video features $\mathcal{F}$ into dense temporal features $\mathcal{D}$ for the bottom hierarchy (Fig. ~\ref{fig:method}-right) and encodes them into downsampled sparse temporal features $\mathcal{S}$ for the upper hierarchies (Fig. ~\ref{fig:method}-left).

To obtain both temporal features, we first partition $\mathcal{F}$ into sliding windows of length $L_w$, producing video segments denoted as $C = [C^{i}]_{i=1,\hdots, |C|}$. Each segment $C^{i}$ is defined by $C^{i}=[f^{s_i+t}]_{t=1,\hdots, L_w}$, where $s_i$ is the start index of each clip, and $C^{i} \in \mathbb{R}^{L_w \times D}$. Dense temporal features $\mathcal{D}^i$ are derived from each segment $C^{i}$ through a linear projection layer, $h_d$, such that $\mathcal{D}^i=h_d(C^i) \in \mathbb{R}^{L_w \times D}$.

To create the sparse features, a two-step process is applied. First, a cross-attention layer (Eq. \ref{eq:cross_attn}) uses segment feature 
$C^i$ as the query and the text feature $Q$ as the key, and outputs text-aligned segment feature $\Tilde{C}^i \in \mathbb{R}^{L_w \times D}$. This cross-attention mechanism aligns the video segment with the text query, enhancing semantic correspondence between the modalities. Next, a self-attention layer (Eq. \ref{eq:ret_attn}) takes the concatenation of a sparse feature $\mathcal{S}^i \in \mathbb{R}^{D}$ and the text-aligned segment feature $\Tilde{C}^i$ as input and condenses the segment into the sparse feature. This sparse feature is a compact, learnable representation similar to the CLS token in BERT~\cite{devlin2019bertpretrainingdeepbidirectional}.

\vspace{-2mm}
\begin{align}
    \Tilde{C}^i &= \operatorname{Cross-Attention}(C^i, Q) \label{eq:cross_attn} \\ 
    A &= \operatorname{Self-Attention}(\textbf{[}\mathcal{S}^i; \Tilde{C}^i\textbf{]}) \label{eq:ret_attn}
\end{align}

For each video segment $C^i$, we compute the sparse temporal feature as $\mathcal{S}^i = {A}_0$, which condenses the segment (e.g., 2-minutes) into a compact embedding (e.g., 768-dimensional), substantially reducing the input context length. Collectively, this process yields the dense temporal features $\mathcal{D} = [\mathcal{D}^{i}]_{i=1,\hdots, |C|}$ and sparse temporal features $\mathcal{S} = [\mathcal{S}^{i}]_{i=1,\hdots, |C|}$ for all segments.
 
\vspace{1mm}
\noindent\textbf{Input for the LLM.}
\label{input_llm}
We construct a video input feature \( [I^{(\ell)}]_{\ell=1, \dots, L} \) with \( L \) hierarchies to capture different temporal scales. As illustrated in Figure \ref{fig:method}, the lowest hierarchical feature, \( I^{(1)} \) is set to the dense features \( \mathcal{D} \) while the higher levels \( [I^{(\ell)}]_{\ell=2, \dots, L} \) use sparse features \( \mathcal{S} \), obtained from the Hierarchical Adapter. We combine this visual input with an instruction prompt for the LLM. We define the instruction as \textit{`` \textless video\textgreater ~ when can we see the \textless event\textgreater ~ happening?''}. Here \textless event\textgreater ~ is replaced by the textual event description. The word embedding layer of the LLM converts this prompt into token embeddings, $[w_1, w_2, \dots, w_M]$. At \textless video\textgreater ~ position, we insert the video features to create the final input prompt, $P^{(\ell)} =[I^{(\ell)}, w_1, \dots , w_M ]$.

\vspace{1mm}
\noindent\textbf{Large Language Model.}
To perform temporal grounding across multiple hierarchies, we utilize a pre-trained language model (e.g., Vicuna~\cite{vicuna2023}) as the temporal grounding decoder, which predicts the event boundaries at each hierarchy. At hierarchy $l$, the LLM receives input, \( P^{(l)} \) and outputs start and end times, \( {\tau^{(l)}} \), for that level in the form of: \textit{`From $s$ to $e$.'}. Here $s$ and $e$ denote the start and end frame indexes of the queried event. If the event is absent from the video segment, the model generates \textit{``Not Present."}. To progressively refine detection, the decoder processes segments features \( P^{(l)} \) containing the predicted boundaries \( {\tau^{(< l)}} \) from the previous hierarchy levels. At the initial level (\( {\tau^{(0)}} \)), no prior boundaries are provided, which requires the model to scan the entire video for initial boundary prediction. The proposed \model\ learns the likelihood of a target event's start and end times \( \tau^{(l)} \) conditioned on the hierarchical video representation $P^{(l)}$ through the following training objective:
\vspace{-2mm}
\begin{align}
p(T^{(l)} | P^{(l)}) = \prod_{k=1}^K p(T^{(\ell)}_{k} | T^{(\ell)}_{< k}, P^{(l)})
\label{eq:model}
\end{align}
\vspace{-2mm}

Here, $T^{(\ell)}_{k}$ denotes the $k^{th}$ language token of the caption, and $T^{(\ell)}_{< k}$ denotes all preceding tokens. 

\begin{figure*}[!t]
    \centering
    \includegraphics[width=1\linewidth]{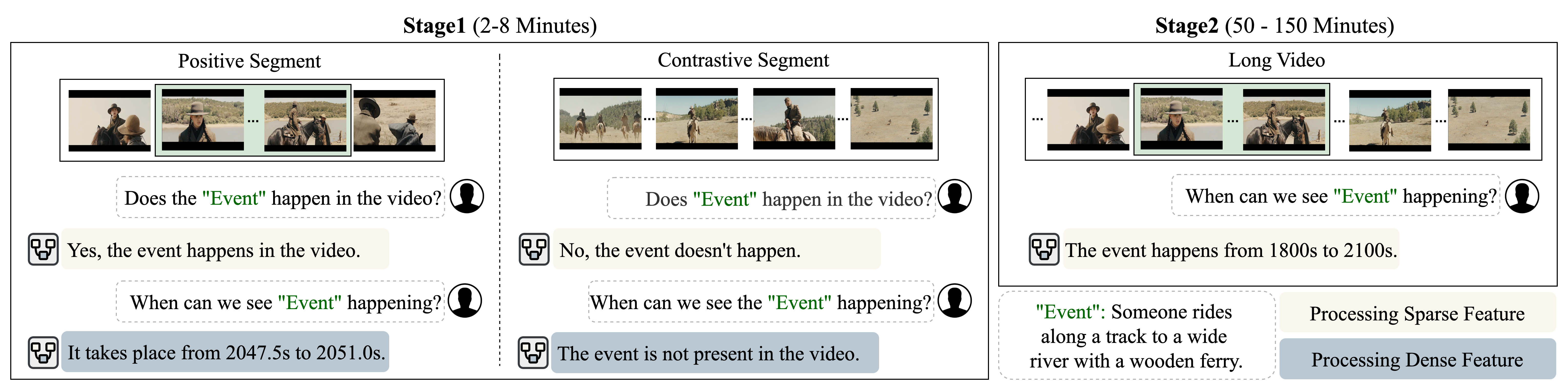}
    \vspace{-6mm}
\caption{\textbf{Progressive Training Method.} Our model is trained progressively: first on short video segments and then on hour-long videos. \textbf{(Left)} In the first stage, the model learns to detect whether an event is present in the input video and, if so, predicts its precise start and endpoints. Sparse features help determine an event's presence, while dense features additionally facilitate exact localization. \textbf{(Right)} In the second stage, we utilize the sparse features learned in Stage 1 to identify event segments within hour-long videos.}
    \vspace{-4mm}
    \label{fig:train}
\end{figure*}

\subsection{Training}
Training a hierarchical video-language model for temporal grounding is challenging, particularly with videos of varying lengths and extensive non-relevant content. To tackle this, we employ a progressive training strategy (Fig. \ref{fig:train}) where the model initially learns to identify events in short video segments before scaling up to hour-long videos. This approach allows the model to first focus on recognizing key events in shorter segments, then apply that understanding effectively to longer, more complex videos.

\vspace{1mm}
\noindent\textbf{Stage 1: Training with Short Segments.} 
\label{sec:stage1}
Traditional VLMs are generally trained with only positive video segments. For example, VTimeLLM~\cite{huang2024vtimellm} trains the model to ground events on video segments that are guaranteed to contain the event (similar to the leftmost subplot in Figure \ref{fig:train}). This approach, however, can lead to overconfidence in the model’s visual predictions~\cite{kostumov2024uncertainty,oh2023towards}. To address overconfidence—a challenge previously tackled in the text domain through contrastive examples~\cite{kapoor2024large}, we introduce contrastive video segments where the target event is intentionally absent (Fig. \ref{fig:train}-middle). In an hour-long video, many video segments are naturally present that are unrelated to the queried event. Including these segments in training helps the model better calibrate its confidence. At first, we use dense features to train the model to predict precise temporal boundaries (e.g., \textit{“From $s$ to $e$.”}) or indicate the absence of events (\textit{“Not Present.”}). During this phase, we fine-tune the LLM component using LoRA. 

After fine-tuning the LLM, we freeze its weights and proceed to fine-tune only the Hierarchical Adapter module to generate sparse temporal features. These sparse features, which are downsampled versions of the original visual data, are crucial for efficient long-video training (see \ref{long_train}). To simplify the training objective for these sparse inputs, we focus on identifying the presence of events rather than locating precise boundaries. We modify the input prompt as \textit{``\textless video\textgreater~Does the \textless event\textgreater~happen in the video? Answer yes or no."} In this phase, the model learns to respond \textit{“Yes.”} for relevant segments or \textit{“No.”} for irrelevant ones. Using this contrastive training strategy, we calibrate the model’s visual confidence and optimize the sparse features needed for effective hour-long video processing.

\vspace{1mm}
\noindent\textbf{Stage 2: Training with Long Videos.}
\label{long_train}
In this stage (Fig. \ref{fig:train}-right), we leverage sparse features to localize relevant segments within hour-long videos. These sparse features correspond to segments typically much longer (e.g., 3 minutes) than the actual target events, allowing the model to efficiently scan and identify broader regions of interest. We use the original prompt defined in Section \ref{input_llm} and keep the weights of the hierarchical perception module fixed, fine-tuning only the same LoRA module used in Stage 1. 

\subsection{Inference with Calibrated Confidence} 
\label{sec:calibration}
Previous methods~\cite{hou2022cone,pan2023scanning} typically used the CLIP~\cite{xue2022clip} similarity score between visual and textual embeddings to rank and select the top-k predictions. In contrast, we rank predictions based on the internal confidence of our LLM. Specifically, we calculate the entropy of the predicted probability distribution for each word generated by the LLM. We do this scoring on the predictions at the bottom hierarchy ($l=0$) where we utilize the dense features $\mathcal{D}$. For $i^{th}$ prediction, let \( p(w | T_{< k}, \mathcal{D}^{(i)}) \) denote the probability of the $k^{th}$ word conditioned on prior words \( T_{< k} \) and corresponding segment feature $\mathcal{D}^{i}$. We calculate the entropy \( H^i_k \) as:
\vspace{-1.5mm}
\[
H^{(i)}_k = - \sum_{w} p(w | T_{< k}, \mathcal{D}^{(i)}) \log p(w | T_{< k}, \mathcal{D}^{(i)})
\]
\noindent Here, $w$ represents each possible word in the vocabulary of our LLM. The overall uncertainty score is obtained by averaging the entropy values across all words in the generated sequence. To convert this into a confidence score ($R^{i}$), we take the inverse of the mean entropy:
\vspace{-1.5mm}
\[
R^{i} = \frac{1}{\frac{1}{K} \sum_{k=1}^{K} H^{i}_k}
\]

\noindent where \( K \) is the total number of generated words. We calculate the confidence score for all the predictions $[R^{i}]_{i=1,\hdots, N_p}$, where $N_p$ is the number of predicted boundaries. This confidence score allows us to rank and select the top-K predictions based on the LLM's internal confidence about its outputs. More details are present in the Supplementary \Cref{sec:sup calibration}.

\section{Experimental Setup}

\subsection{\model\  Baselines}
Long video temporal grounding remains largely uncharted territory for vision-large language models, leaving a lack of established baselines for meaningful comparison. To address this, we present the following video-language baselines that we have adapted specifically for this task.
\begin{enumerate}
    \item \textbf{VTimeLLM}~\cite{huang2024vtimellm}. A fully fine-tuned baseline that takes in hour-long video as input and produces temporal boundary localization. 
    \item \textbf{VTimeLLM + CONE}~\cite{huang2024vtimellm,hou2022cone}. This baseline is fully fine-tuned on shorter video segments as ours. We employ CONE’s fine-grained ranking method to select the top-k predictions across all segments. First, mean pooling aggregates the CLS features from CLIP~\cite{xue2022clip} across all frames to create an average frame representation. The similarity score is then computed with a dot product between this averaged frame feature and the query text CLS feature from CLIP.
\end{enumerate}

\begin{table*}[!t]
    \begin{adjustbox}{width=1\textwidth}
    \small
    \begin{tabular}{l|ccccccc|cccccc|c}
    \toprule
    \multirow{2}{*}{\textbf{Model}} & 
    \multicolumn{7}{c|}{\textbf{MAD} \cite{soldan2022mad}} & \multicolumn{5}{c}{\textbf{VidChapters-7m} \cite{yang2023vidchapters7mvideochaptersscale}} && \multirow{2}{*}{\textbf{Average↑}} \\
    \cmidrule{2-8} \cmidrule{9-14} 
    & \textbf{R1@.1} & \textbf{R5@.1} & \textbf{R1@.3} & \textbf{R5@.3} & \textbf{R1@.5} & \textbf{R5@.5} & \textbf{Avg.↑} & & \textbf{R1@.3} & \textbf{R1@.5} & \textbf{R1@.7} & \textbf{R1@.9} & \textbf{Avg.↑} & \\
    \midrule
    M-Guide~\cite{barrios2023localizing} & 9.3 & 18.9 & 4.6 & 13.1 & 2.2 & 7.4 & 9.3 & & \textcolor{gray}{-} & \textcolor{gray}{-} & \textcolor{gray}{-} & \textcolor{gray}{-} & \textcolor{gray}{-} & \textcolor{gray}{-} \\
    CONE~\cite{hou2022cone} & 8.9 & 20.5 & 6.9 & 16.1 & 4.1 & 9.6 & 11.0 & & \textcolor{gray}{-} & \textcolor{gray}{-} & \textcolor{gray}{-} & \textcolor{gray}{-} & \textcolor{gray}{-} & \textcolor{gray}{-} \\
    SOONet~\cite{pan2023scanning} & 11.3 & 23.2 & 9.0 & 19.6 & 5.3 & \underline{13.1} & 13.6 & & \textcolor{gray}{-} & \textcolor{gray}{-} & \textcolor{gray}{-} & \textcolor{gray}{-} & \textcolor{gray}{-} & \textcolor{gray}{-} \\
    SnAG~\cite{mu2024snag} & 10.3 & 24.4 & 8.5 & \underline{20.6} & 5.5 & \textbf{13.7} & {13.8} & & \textcolor{gray}{-} & \textcolor{gray}{-} & \textcolor{gray}{-} & \textcolor{gray}{-} & \textcolor{gray}{-} & \textcolor{gray}{-} \\
    RGNet~\cite{hannan2023rgnet} & {12.4} & {25.1} & {9.5} & 18.7 & {5.6} & 10.9 & 13.7 & & \textcolor{gray}{-} & \textcolor{gray}{-} & \textcolor{gray}{-} & \textcolor{gray}{-} & \textcolor{gray}{-} & \textcolor{gray}{-} \\
    BERT~\cite{devlin2019bertpretrainingdeepbidirectional} & \textcolor{gray}{-} & \textcolor{gray}{-} & \textcolor{gray}{-} & \textcolor{gray}{-} & \textcolor{gray}{-} & \textcolor{gray}{-} & \textcolor{gray}{-} & & 0.6 & 0.3 & 0.1 & 0.0 & 0.3 & 0.3 \\
    VTimeLLM$^{\ast}$~\cite{zhang2020learning} & 1.4 & 3.1 & 1.3 & 2.5 & 0.6 & 1.1 & 1.7 & & 10.6 & 4.1 & 1.6 & 0.2 & 4.1 & 2.9 \\
    CLIP~\cite{radford2021learning} & 6.6 & 15.1 & 3.1 & 9.9 & 1.5 & 5.4 & 6.9 & & 10.7 & 5.2 & 2.3 & 0.5 & 4.7 & 5.8 \\
    M-DETR~\cite{lei2021detecting} & 3.6 & 13.0 & 2.8 & 9.9 & 1.7 & 5.6 & 6.1 & & \textbf{37.4} & \underline{27.3} & \underline{17.6} & \underline{6.4} & \underline{22.1} & \underline{14.1} \\
    \textbf{Ours}$^{\dagger}$ & \textbf{17.3} & \textbf{31.4} & \textbf{12.7} & \textbf{23.5} & \textbf{6.7} & \underline{13.1} & \textbf{17.5} & & \textcolor{gray}{-} & \textcolor{gray}{-} & \textcolor{gray}{-} & \textcolor{gray}{-} & \textcolor{gray}{-} & \textcolor{gray}{-}  \\
    \textbf{Ours} & \underline{15.0} & \underline{25.1} & \underline{11.0} & {18.8} & \underline{5.8} & {10.5} & \underline{14.4} & & \underline{33.8} & \textbf{27.4} & \textbf{21.8} & \textbf{15.2} & \textbf{24.6} & \textbf{19.5} \\
    \bottomrule
    \end{tabular}
    \end{adjustbox}
    \vspace{-2mm}
    \caption{\textbf{Main Results on the MAD and VidChapters-7M Datasets.} The best scores are highlighted in \textbf{bold}, while the second-best scores are \underline{underlined}. ReVisionLLM demonstrates state-of-the-art performance across both datasets. $^{\ast}$This paper trains VTimeLLM on the datasets and uses CONE\cite{hou2022cone} ranking method. $^{\dagger}$\model-I variant processes more frames and achieves higher accuracy.}
    \vspace{-6mm}
    \label{tab:sota}
\end{table*}

\subsection{\model\  Variations}
\begin{enumerate}
    \item \textbf{\model .} Our default model begins by processing hour-long video segments at the top of the hierarchy, then moves down to shorter segments. In this setup, we train two LoRAs within the LLM: one for the bottom hierarchy and another for the higher levels. This approach is highly efficient for inference, as it reduces the number of input frames processed by the LLM.
    \item \textbf{\model-U.} This is a unified model across all hierarchies with shared weights. With significantly fewer trainable parameters than the default model, it is more efficient to train.
    \item \textbf{\model-I.} In this variant, the LLM operates in reverse order, starting from the bottom of the hierarchy and progressively increasing the video length. Like the default model, it uses two LoRAs but requires processing all video frames, trading efficiency for improved performance with added computation.
    \item \textbf{\model-(U+I).} This variant has a unified architecture like \model-U and starts processing from the bottom hierarchy like \model-I. 

\end{enumerate}
\subsubsection{Datasets and Metrics}
\noindent\textbf{MAD Dataset}~\cite{soldan2022mad}. This large-scale dataset comprises approximately 1,200 hours of full-length movies, featuring 384,000 natural language queries linked to specific moments within the videos. On average, each video is around 110 minutes long, while the moments are brief—just 4.1 seconds on average—making the moment-to-video ratio very low. This disparity poses a substantial challenge for accurate temporal grounding.

\vspace{1mm}
\noindent\textbf{VidChapters-7M} ~\cite{yang2023vidchapters7mvideochaptersscale}. This is a large-scale, user-annotated dataset with over 7 million chapters across 817,000 videos, with the longest 12 hours. Each video includes 2 to 30 chapters with durations from 1 second to 10 minutes, making it a challenging dataset for temporal localization due to its length and variety.

\vspace{1mm}
\noindent\textbf{Evaluation Metrics.} Following prior work \cite{soldan2022mad, hou2022cone}, we use Recall@$k$ at IoU=$\theta$ ({Rk@$\theta$}) as the primary grounding metric. This measures the proportion of test samples where at least one of the top-$k$ predictions achieves an Intersection over Union (IoU) greater than $\theta$ with the ground truth. Additionally, to assess generalization to Text-to-Video retrieval, we use the standard Recall at Rank k (R@k) metric~\cite{gorti2022xpoolcrossmodallanguagevideoattention}, which measures the percentage of ground truth video in the top-k retrieved ones.

\vspace{1mm}
\noindent\textbf{Implementation Details.}  In our approach, we utilize the 7B version of Vicuna v1.5~\cite{vicuna2023} as the Large Language Model. Training is conducted on the using a total batch size of 128 across 8 A100 GPUs. For optimization, we use AdamW~\cite{loshchilov2019decoupledweightdecayregularization} with a cosine learning rate decay and an initial warm-up phase. During the adapter training stage, we run 1 epoch with a learning rate of $1\times 10^{-3}$. In the subsequent hierarchical stage, we train for 5 epochs for MAD and 1 epoch for VidChapters-7M, with a learning rate of $1\times 10^{-4}$. LoRA settings include parameters $r=64$ and $alpha=128$. Please refer to the Supplementary \Cref{sec:sup implementation} for additional implementation details.

\section{Results}
In this section, we present our performance against previous methods, detailed ablation studies, qualitative results, and generalizations of text-to-video retrieval tasks.

\vspace{1mm}
\noindent\textbf{Main Results on the MAD Dataset}~\cite{soldan2022mad}.
\model\ sets a new state-of-the-art on the MAD dataset, outperforming prior models in temporal grounding (Tab. \ref{tab:sota}). It surpasses the previous best method, RGNet~\cite{hannan2023rgnet}, by +2.6\% in R1@.1 and +1.5\% in R1@.3, achieving competitive scores across other metrics. Our \model-I variant outperforms RGNet by an even larger margin, +4.9\% in R1@.1 and +6.3\% in R1@.3. As the moments-to-video ratio is extremely low in this dataset, it requires fine-grained event understanding. Our model’s recursive architecture effectively narrows the search for relevant segments, handling these challenges well. While existing methods relying on heuristic CLIP~\cite{xue2022clip} ranking struggle with numerous false detections (evident by low R1 score across all thresholds) in this dataset, \model\ reliance on LLM's internal confidence reduces such errors.

\vspace{1mm}
\noindent\textbf{Main Results on VidChapters-7M Dataset}~\cite{yang2023vidchapters7mvideochaptersscale}. On the VidChapters-7M dataset, \model\ sets a new state-of-the-art (Table \ref{tab:sota}), significantly outperforming the previous best model, M-DETR~\cite{lei2021detecting}, particularly at higher IoU thresholds (+4.2\% in R1@.7 and +8.8\% in R1@.9) and showing strong results across other metrics. This precision at stricter thresholds demonstrates the superior ability of \model\ to localize events accurately. The dataset includes a diverse range of YouTube tutorial videos with user-queried steps, underscoring our model’s advancements in improving video content search for online platforms across short clips to extended videos up to 12 hours.

\subsection{Ablation Studies} We present ablation studies on each module, model variants, video length, and the number of hierarchies by experimenting on the MAD dataset~\cite{soldan2022mad}.

\begin{table}[!ht]
\centering
\scriptsize
\setlength{\tabcolsep}{3.5pt} 
\renewcommand{\arraystretch}{1.2} 
\begin{tabular}{l|cccc}
\toprule
\textbf{Modules} & \textbf{R1@.1↑} & \textbf{R5@.1↑} & \textbf{R1@.3↑} & \textbf{R5@.3↑} \\
\midrule
\textit{Baseline:} VTimeLLM~\cite{huang2024vtimellm} & 0.0 & 0.0 & 0.0 & 0.0 \\ 
\hspace{3mm}(+) CONE~\cite{hou2022cone} & 1.4 & 2.4 & 1.3 & 2.5 \\ 
\hspace{3mm}(+) Contrastive Segment & 4.8 & 6.7 & 4.2 & 7.2 \\ 
\hspace{3mm}(+) Calibration \textcolor{gray}{(-) CONE}& 8.4 & 12.7 & 6.6 & 8.9 \\ 
\hspace{3mm}(+) Recursive Process$^{\ast}$ & \textbf{15.0} & \textbf{25.1} & \textbf{11.0} & \textbf{18.8}  \\ 
\bottomrule
\end{tabular}
\vspace{-2mm}
\caption{\textbf{Cumulative Ablation on Proposed Modules.} Each of our proposed modules contributes to a significant improvement in grounding capability, with the recursive process achieving the highest gains. $^{\ast}$Indicates the \model\ model.}
\vspace{-4mm}
\label{tab:ablation_modules}
\end{table}

\vspace{1mm}
\noindent\textbf{Modules.}
Table \ref{tab:ablation_modules} highlights the unique contributions of each module in our model for temporal grounding on long videos. We start with the baseline VTimeLLM~\cite{huang2024vtimellm} model, trained on the MAD dataset, which scores zero across all recall metrics due to uniform sampling of 100 frames from hours-long videos, causing a complete loss of temporal details. Next, we train and test the VTimeLLM model on shorter video segments and apply CONE’s ranking strategy~\cite{hou2022cone} for final predictions. This single level of hierarchical processing yields modest improvements (e.g., R1@.1 of 1.4\% and R5@.1 of 2.4\%). 
 
The addition of the Contrastive Segments marks a notable improvement, raising R1@.1 to 4.8\% and R5@.1 to 6.7\%; by allowing the model to identify absent segments (e.g., ``Not Present"), it narrows the temporal search space and enhances segment selection. Replacing CONE with our Grounding LLM’s confidence-based ranking further boosts results (e.g., R1@.1 of 8.4\% and R5@.1 of 12.7\%), as training on positive and contrastive segments in Stage 1 improves the LLM’s calibration, aligning confidence with prediction accuracy and enhancing ranking effectiveness. Finally, the recursive process achieves the highest performance gains, with R1@.1 of 15.0\% and R5@.1 of 25.1\%, by progressively refining the temporal focus. Each module contributes critically, with the Recursive Process module delivering the largest gains.

\begin{table}[!h]
\centering
\scriptsize
\renewcommand{\arraystretch}{1.2} 
\setlength{\tabcolsep}{.75mm} 
\begin{tabular}{l|c|c|cc}
\toprule
\textbf{Model} & \textbf{Train Params$\downarrow$} & \textbf{Input Frames$\downarrow$} & \textbf{R1@.1↑} & \textbf{R5@.1↑}  \\
\midrule
{VTimeLLM}+CONE & 363M& 100\%&1.4 & 2.4  \\ \midrule
\textbf{\model} & 363M& \textbf{57\%} & 15.0 & 25.1  \\
\model-U & \textbf{159M}& 58\% & 14.4 & 24.7  \\
\model-(U+I) & \textbf{159M}& 100\% & 16.7 & 31.0  \\
\model-I & 363M& 100\% & \textbf{17.4} & \textbf{31.4} \\
\bottomrule
\end{tabular}
\vspace{-2mm}
\caption{\textbf{Ablation on Model Variants.}  We perform well with fewer frames and trainable parameters than the baseline, which struggles to solve the task effectively. Processing additional frames and more parameters further improves our performance.}    
\vspace{-4mm}
\label{tab:ablation_variants}
\end{table}

\vspace{1mm}
\noindent\textbf{Model Variants.}
Table~\ref{tab:ablation_variants} compares \model\ variants to assess the effects of hierarchical processing order, percentage of video input frames for the LLM, and parameter sharing. The baseline (\textit{VTimeLLM}+CONE) performs poorly (R1@.1 = 1.4\%), despite processing 100\% of video frames as input to the LLM, underscoring the limitations of non-recursive approaches. In contrast, our default \model\ processes recursively from top to bottom, balancing accuracy and efficiency, achieving R1@.1 of 15.0\% and R5@.1 of 25.1\% with processing only 57\% of input frames. The \model-U variant enhances training efficiency by sharing weights across all hierarchies, resulting in a slight performance reduction (R1@.1 = 14.4\%, R5@.1 = 24.7\%) while using fewer trainable parameters (363M vs. 159M). The \model-I variant reaches the highest accuracy (R1@.1 = 17.4\%, R5@.1 = 31.4\%) by processing in reverse hierarchical order (bottom to top), with more input frames. \model-(U+I) combines reverse processing with parameter sharing, balancing training efficiency, and high performance (R1@.1 = 16.7\%, R5@.1 = 31.0\%). Overall, the default \model~offers strong accuracy with high frame efficiency, while \model-I provides peak accuracy with increased input frames to the LLM.

\begin{figure}[!h]
    \centering
    \includegraphics[width=.5\linewidth]{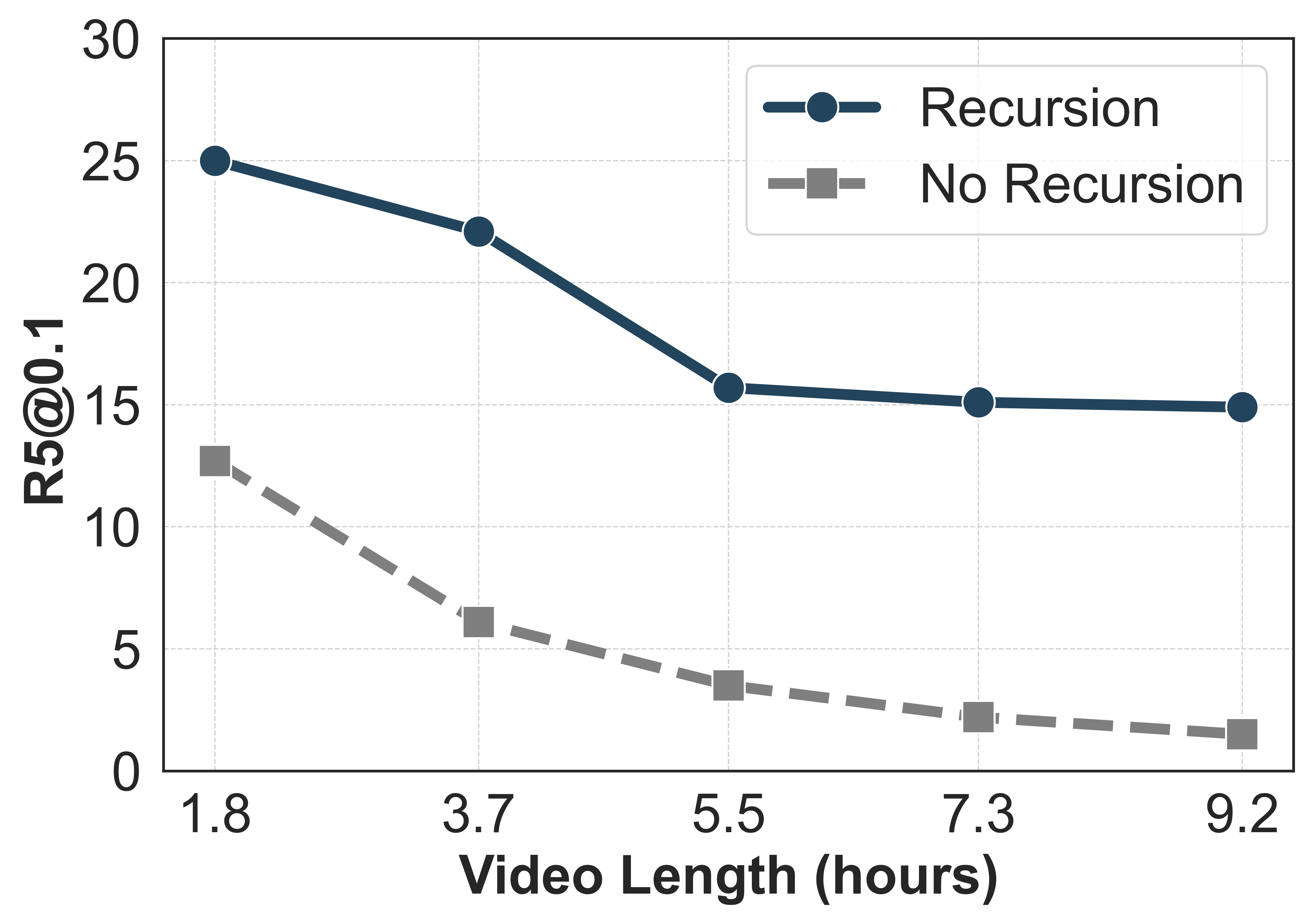} 
    \vspace{-2mm}
    \caption{\textbf{Ablation on Video Length.} Our recursive approach maintains strong performance even with videos up to 10 hours long, while the baseline method fails entirely in these cases.}
    \vspace{-4mm}
    \label{fig:performance_plot}
\end{figure}

\begin{figure*}[!t]
    \centering
    \includegraphics[width=.9\linewidth]{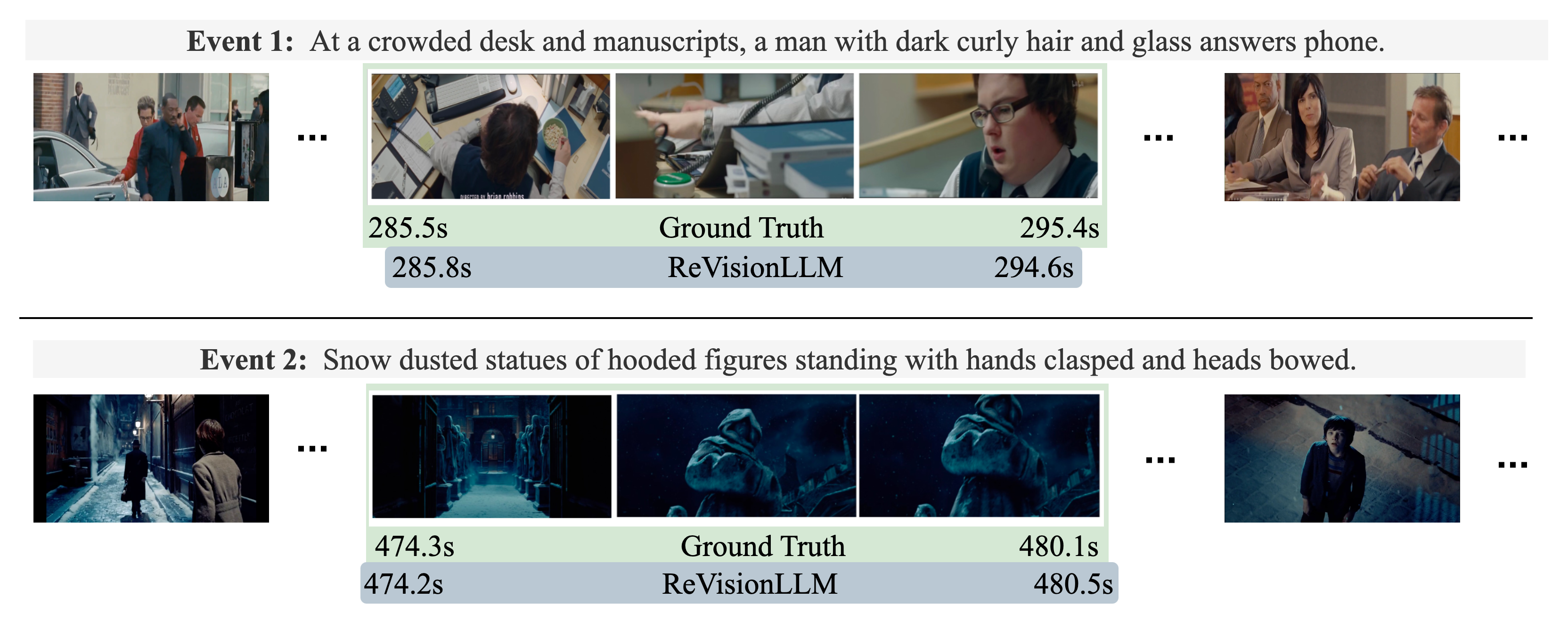}
    \vspace{-4mm}
    \caption{\textbf{Qualitative results on MAD.} \model\ accurately locates precise event boundaries that involve intricate actions (top) and complex visual details (bottom) within hour-long movies. In contrast, our VLM baseline fails entirely to capture these events.}
    \vspace{-6mm}
    \label{fig:qual}
\end{figure*}

\vspace{1mm}
\noindent\textbf{Video Length.} Fig. \ref{fig:performance_plot} demonstrates ReVisionLLM's robustness in handling long videos. We extend the videos by repeating them multiple times to create longer sequences, ensuring that the ground truth moment appears only once within the extended video. While the method without recursion fails for 10-hour videos, recursive video processing maintains strong performance. The slight decrease in performance for longer videos reflects the inherent challenges of temporal grounding in extended content rather than a limitation of the model. Beyond approximately five hours, performance stabilizes, as the increasing diversity of scenes has minimal impact on event localization.

\vspace{1mm}
\noindent\textbf{Number of Hierarchies.}
Table \ref{tab:hierarchy_levels_ablation} shows the impact of the number of hierarchies. Without hierarchy, treating the entire video as a single unit prevents the model from capturing event boundaries, highlighting the limitations of current non-recursive VLMs in long video grounding. With one hierarchy, we segment the video and aggregate predictions with calibrated confidence, showing some improvement but still struggling with high false positives. With 2 and 3 hierarchies, we progressively filter out irrelevant regions and revise predictions recursively, resulting in higher accuracy.
\begin{table}[!h]
\centering
\scriptsize
\renewcommand{\arraystretch}{1.15} 
\begin{tabular}{c|cccc}
\toprule
\textbf{Hierarchies} & \textbf{R1@.1↑} & \textbf{R5@.1↑} & \textbf{R1@.3↑} & \textbf{R5@.3↑} \\
\midrule
0 & 0.0 & 0.0 & 0.0 & 0.0 \\
1 & 8.4 & 12.7 & 6.6 & 8.9 \\
2 & 11.9 & 17.5 & 8.7 & 13.2 \\
3 & \textbf{15.0} & \textbf{25.1} & \textbf{11.0} & \textbf{18.8} \\
\bottomrule
\end{tabular}
\vspace{-2mm}
\caption{\textbf{Ablation on number of Hierarchies.} The model’s performance improves with the number of hierarchies in the recursive structure, becoming more effective with each additional level. Without this hierarchical approach, the model fails on the task.}
\vspace{-6mm}
\label{tab:hierarchy_levels_ablation}
\end{table}

\subsection{Qualitative Results on MAD Dataset} In Figure \ref{fig:qual}, we demonstrate two examples of events accurately localized by \model. In the first example, frequent scenes of office work closely resemble the queried event, but \model\ successfully identifies the specific instance where a person with distinct attributes answers a phone—a task requiring detailed comprehension of both appearance and action. The second example tests the model's ability to identify a complex visual description spanning 5.8 seconds within a 2-hour movie, highlighting its effectiveness in locating subtle differences within visually similar footage. More qualitative results are included in the Supplementary \Cref{sec:sup qualitative}.

\subsection{Generalization: Text-to-Video Retrieval}
Text-to-video retrieval is the task of identifying the video corresponding to a textual event description from a large set of different videos. We solve this task with our \model\ by concatenating all the videos into a single hour-long video and applying our model to locate the index of the predicted video. It outperforms previous state-of-the-art models on the MSRVTT~\cite{msrvtt} dataset, with improvements of +2.2\% in R@5 and +0.6\% in R@10, while achieving competitive R@1 performance. These results demonstrate \model's understanding of video-text correspondence, positioning it as a strong model for general multi-modal retrieval tasks, including large-scale video searches.
More details are provided in the Supplementary \Cref{sec:sup generalization}.

\vspace{-.5mm}
\begin{table}[!h]
\centering
\label{tab:retrieval_results}
\scriptsize
\begin{tabular}{l|ccc}
\toprule
\textbf{Method}
& \textbf{R@1 $\uparrow$} & \textbf{R@5 $\uparrow$} & \textbf{R@10 $\uparrow$} \\
\midrule
X-Pool \cite{gorti2022xpoolcrossmodallanguagevideoattention}  & 46.9 & 72.8 & 82.2 \\
DiffusionRet \cite{jin2023diffusionretgenerativetextvideoretrieval} & 49.0 & 75.2 & 82.7 \\
UATVR \cite{fang2023uatvruncertaintyadaptivetextvideoretrieval} & 47.5 & 73.9 & 83.5 \\
TEFAL \cite{ibrahimi2023audioenhancedtexttovideoretrievalusing} & 49.4 & {75.9} & 83.9 \\
CLIP-ViP \cite{xue2023clipvipadaptingpretrainedimagetext} & 50.1 & 74.8 & 84.6 \\
T-MASS \cite{wang2024textmassmodelingstochastic} & \textbf{50.2} & 75.3 & {85.1} \\
\textbf{Ours} & 49.1 & \textbf{77.5} & \textbf{85.7} \\
\bottomrule
\end{tabular}
\vspace{-2mm}
\caption{\textbf{\model's Generalization.} Our model generalizes well to the Text-to-Video retrieval task and performs competitively with state-of-the-art models on the MSRVTT~\cite{msrvtt} dataset.}
\vspace{-6mm}
\label{tab:retrieval_results}
\end{table}

\section{Conclusion and Future Work}
We introduce \model, the first VLM specifically designed with a recursive structure for temporal event grounding in \textbf{hour-long videos}. Its recursive architecture effectively can locate events within extensive videos and establishes a new state-of-the-art, outperforming specialized models. Future work could focus on integrating audio for better event comprehension and expanding the capabilities to handle even longer videos spanning multiple days. 
{
    \small
    \bibliographystyle{ieeenat_fullname}
    \bibliography{main}
}
\clearpage
\setcounter{page}{1}
\maketitlesupplementary
\setcounter{section}{0}
\setcounter{figure}{0}
\setcounter{table}{0}

\renewcommand{\thesection}{S\arabic{section}}
\renewcommand{\thetable}{S\arabic{table}}
\renewcommand{\thefigure}{S\arabic{figure}}

Our supplementary materials contain \Cref{sec:sup implementation}: Additional Implementation Details, \Cref{sec:sup calibration}: Calibration Confidence, \Cref{sec:sup generalization}: Generalization: Text-to-Video Retrieval, and \Cref{sec:sup qualitative}: Aditional Qualitative Results.

\section{Additional Implementation Details}
\label{sec:sup implementation}
\begin{figure}
    \centering
    \includegraphics[width=1\linewidth]{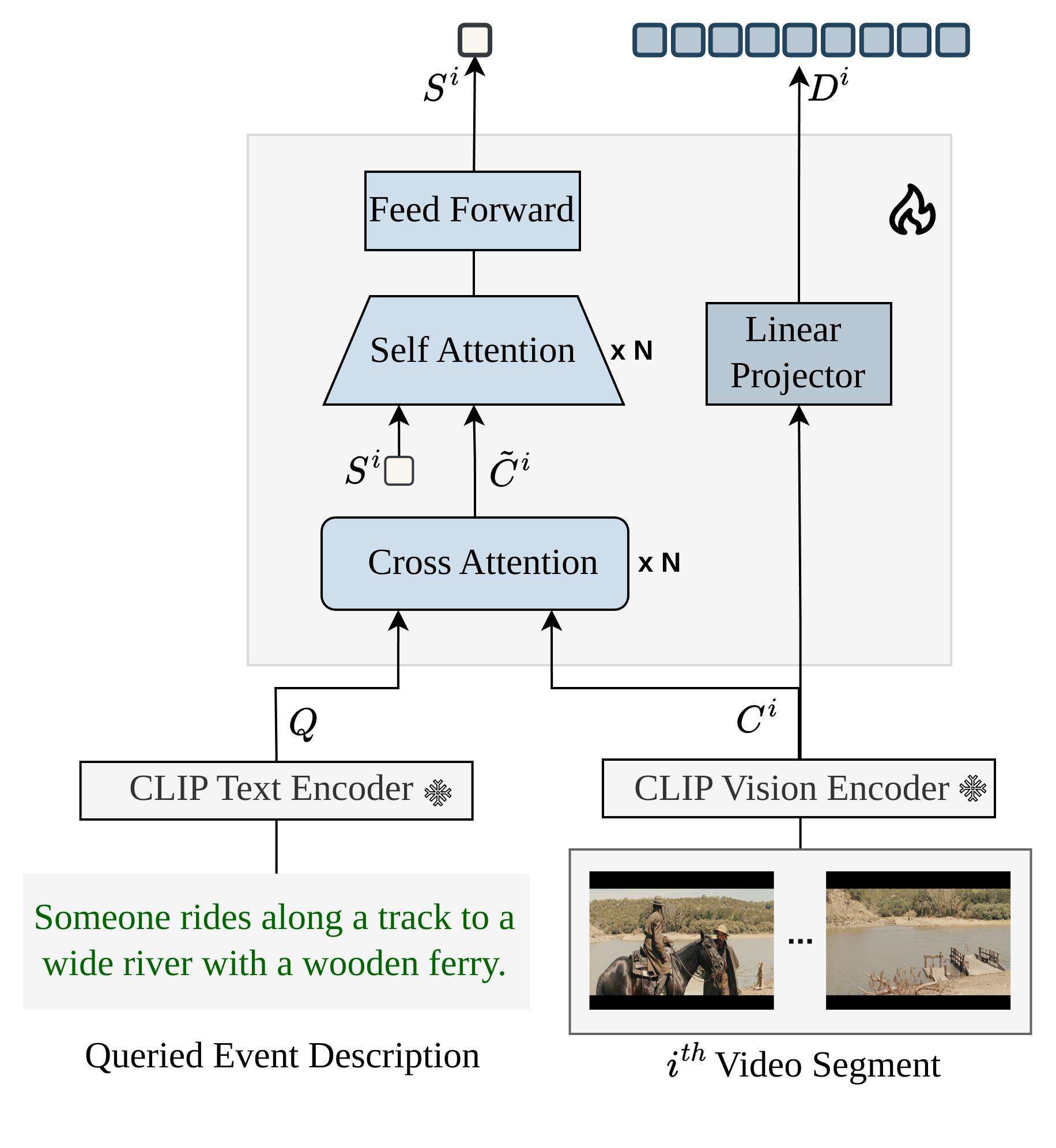}
    \caption{\textbf{Hierarchical Adapter} processes the features extracted by the multimodal encoder, using both the video segments and the textual description of the queried event as inputs. It generates two types of temporal features: sparse and dense. Sparse features are computed through a combination of cross-attention, self-attention, and a feed-forward network, while dense features are generated using a linear projection layer.}
    \label{fig:sup hier_adapter}
\end{figure}

\noindent\textbf{Multimodal Encoder.}
To handle multimodal inputs (refer to Fig.~\ref{fig:sup hier_adapter}), we utilize the Frozen CLIP L/14 model~\cite{xue2022clip}, a 24-layer transformer pretrained on extensive image-text pairs using a contrastive learning objective~\cite{zhao2023learning}. The vision encoder processes inputs of size $224 \times 224$, producing both spatial tokens and a global CLS token. For computational efficiency, we use only the CLS token to represent each frame in long videos. Similarly, the CLIP text encoder, a 12-layer transformer, extracts feature representations for the queried event from the input text.

\noindent\textbf{Hierarchical Adapter.}
As illustrated in Fig.~\ref{fig:sup hier_adapter}, the Hierarchical Adapter processes the $i^{\text{th}}$ video segment $C^i$ to generate both sparse ($S^i$) and dense ($D^i$) features. For the MAD dataset, video features are divided into sliding windows of $L_w = 125$ seconds, while for the VidChapters-7M dataset, the window length is $L_w = 500$ seconds. From each segment, $250$ frames are uniformly sampled and used as input to the hierarchical adapter. 

Sparse features $S^i$ are computed using a combination of cross-attention and self-attention mechanisms, each implemented with two layers ($N=2$). This lightweight design ensures minimal computational overhead compared to the 24 transformer layers of the original CLIP Vision Encoder. Dense features $D^i$ are derived by projecting the CLIP-encoded frame features (dimension 768) into the embedding space of the Large Language Model (dimension 4096)~\cite{vicuna2023} using a linear transformation.

\noindent\textbf{Large Language Model.}
We utilize a pre-trained Vicuna-7B~\cite{vicuna2023} model to ground queried events using the adapted visual features. Built upon LLaMA~\cite{touvron2023llama}, this model consists of 32 transformer layers and has been fine-tuned on 70K user-shared conversations from ShareGPT~\cite{ShareGPT59}. 

To enhance training efficiency, we adopt Low-Rank Adaptation (LoRA)~\cite{hu2021lora}, a method commonly used in recent works~\cite{huang2024vtimellm,ren2024timechat}. LoRA allows us to fine-tune the model without modifying its core weights by introducing lightweight, trainable modules. This significantly reduces computational overhead while retaining the model's flexibility. For our setup, we configure LoRA with a rank of \(r = 64\) and a scaling factor of \(\alpha = 128\).

\noindent\textbf{Training on \model\ Model}
We begin by pretraining the Linear Projector (Fig. \ref{fig:sup hier_adapter}) using the LCS-558K dataset from LLaVA~\cite{liu2023llava}. This step aligns the CLS token from the CLIP Vision Encoder with the LLM's embedding space. The projector is trained for 1 epoch with a batch size of 128 and a learning rate of $1 \times 10^{-3}$.

Following pretraining, we implement a two-stage training pipeline for \model, maintaining a consistent learning rate of $1 \times 10^{-4}$. We use the AdamW optimizer~\cite{loshchilov2019decoupledweightdecayregularization} with a warmup ratio 0.03 and a cosine scheduling strategy.

In the first stage, the Linear Projector is frozen, and the LLM is fine-tuned using LoRA on dense features, focusing on the lowest hierarchy level. This stage employs a batch size of 128, spanning 5 epochs for the MAD dataset and 1 epoch for the VidChapters-7M dataset. Sparse temporal features are introduced for upper hierarchies to reduce the LLM's visual input size. To enable sparse feature generation, we freeze the LoRA module and fine-tune the Cross-Attention, Self-Attention, and Feed-Forward layers of the Hierarchical Adapter, using a batch size of 32 for 1 epoch.

The training in this stage incorporates contrastive video segments where the queried event is absent. These segments are randomly sampled from hour-long videos and do not overlap with the temporal boundaries of the ground truth event. By selecting contrastive segments from the same video, the model is trained to handle challenging inference scenarios, where it must distinguish the queried event from visually and contextually similar scenes within the video.

In the second stage, all components of the Hierarchical Adapter are frozen, and a new LoRA module is fine-tuned for long-video processing on the Stage 2 objective. This stage employs a batch size of 8 and runs for 2 epochs. Two separate LoRA modules are utilized: one optimized for short video training and another adapted for long video processing.

\noindent\textbf{Training on \model-U Model}
The unified model variant differs from our default model only in its training methodology. Training a unified model with shared parameters across all hierarchical levels poses notable challenges. To address this, we adopt an enhanced two-stage strategy. The first stage, including pretraining, is similar to the procedure used in the \model~framework. In the second stage, however, we introduce a dual-training approach, where the \model-U framework is simultaneously trained on both short video clips and hour-long videos. This approach reduces the risk of catastrophic forgetting, ensuring the retention of short-segment representations.

A key challenge arises from the significant differences between short-segment and long-video data. Short segments utilize dense temporal features, while long videos rely on sparse temporal representations, such as CLS features. Additionally, short-segment training involves only video features, whereas long-video descriptions require both video and text features as inputs. To reconcile these differences, we implement an alternating batching strategy. During training, batches of short segments and long videos are alternately sampled, enabling the model to learn effectively from both data types.

This alternating training strategy not only mitigates catastrophic forgetting but also facilitates the successful training of the \model-U framework, which maintains shared parameters across all hierarchical levels. For \model-U, we employ the same hyperparameters as those used in \model~, including learning rate, batch size, training epochs, optimizer, and scheduler (as detailed in the previous section).

\noindent\textbf{Inference for \model.} During inference, video segments are created using a sliding window approach. For the MAD dataset, each segment spans 125 seconds with a stride of 25 seconds, while for the VidChapters-7M dataset, segments are 500 seconds long with a stride of 100 seconds. From each segment, 250 frames are uniformly sampled. Sparse temporal features are then extracted using the Hierarchical Adapter and provided as input to the LLM to identify relevant video segments.

In our implementation, we employ two hierarchies with long videos. However, it can be extended to more levels based on video length. At the top level, 100 video segments (approx. 150 minutes) are processed simultaneously, while the second level processes 33 segments (approx. 50 minutes) simultaneously. Both hierarchies identify regions of interest, refined at the lowest hierarchical level. In this final hierarchy, all 250 dense temporal features from the selected segments are processed to pinpoint the precise event boundaries.

\noindent\textbf{Inference for \model-I.} The inverse model variant differs from our default model only in the inference method. In this variant, the inference begins at the lowest hierarchical level. All video segments are processed together in a single input batch, with their dense temporal features fed into the LLM. The LLM predicts temporal boundaries for multiple segments, often resulting in a high number of false positives. To mitigate this, the false positives are recursively passed through the second and third hierarchical levels, where they are filtered out. These upper levels retain only the most confident predictions, reducing errors. Finally, the confidence scores from the higher hierarchies are used to adjust and normalize the scores of the initial predictions, improving the overall accuracy of the model. We will release the code and pre-trained models for further use.

\section{Calibration Confidence} 
\label{sec:sup calibration}
Accurate calibration of our model's confidence is crucial for minimizing false positives and improving the overall effectiveness of our approach. To evaluate the impact of our training strategy on model calibration, we compare our method's performance to the baseline VTimeLLM~\cite{huang2024vtimellm}. A model is considered well-calibrated if its confidence scores match the actual proportion of correct predictions. Calibration is typically measured using the Expected Calibration Error (ECE)~\citep{Naeini2015ObtainingWC}, which quantifies the difference between predicted probabilities and observed outcomes by dividing the predicted confidence into discrete bins. A lower ECE value indicates better calibration, with an ECE of 0 representing perfect calibration.

For a dataset of $N$ video segments and $B$ evenly spaced bins $b_j$, the ECE is computed as follows:
\[
    \widehat{\text{ECE}} = \sum_{j=1}^B \frac{\lvert b_j \rvert}{N} \left\lvert \mathrm{conf}(b_j) - \mathrm{acc}(b_j) \right\rvert
\]
where $\mathrm{conf}(b_j)$ is the average confidence of samples in bin $b_j$, $\mathrm{acc}(b_j)$ is the accuracy of predictions in bin $b_j$, and $\lvert b_j \rvert$ is the number of samples in bin $b_j$. In our experiments, $\mathrm{conf}(b_j)$ corresponds to the confidence score ($R^i$) defined in Section 3.4 of the main paper.
A prediction is considered correct if the Intersection over Union (IoU) exceeds a threshold $\tau_{\text{iou}} \in \{0.1, 0.3, 0.5\}$.

By comparing ECE values across models, we can assess the effectiveness of our training strategy in improving calibration and generating more reliable confidence estimates.

\begin{table}[h]
\centering
\renewcommand{\arraystretch}{1.2}
\setlength{\tabcolsep}{8pt}
\begin{tabular}{l|ccc}
\toprule
\multirow{2}{*}{\textbf{Model}} & \multicolumn{3}{c}{\textbf{ECE @ IoU Thresholds ($\tau_{\text{IoU}}$)}} \\ 
\cmidrule{2-4}
                                & {$\tau = 0.1 \downarrow$} & {$\tau = 0.3 \downarrow$} & {$\tau = 0.5 \downarrow$} \\ 
\midrule
VTimeLLM$^{\ast}$               & 0.6231                   & 0.6233                   & 0.6237                   \\ 
\model\                         & \textbf{0.4614}          & \textbf{0.4698}          & \textbf{0.4791}          \\ 
\bottomrule
\end{tabular}
\caption{\textbf{Expected Calibration Error (ECE)} comparison between $^{\ast}$Baseline (VTimeLLM+CONE) and Our Model across IoU thresholds ($\tau_{\text{IoU}}$). Our model demonstrates better calibration of confidence compared to the baseline across all IoU thresholds. Lower values indicate superior calibration performance.}
\label{tab:ece_comparison}
\end{table}

Table \ref{tab:ece_comparison} presents a comparison of Expected Calibration Error (ECE) values between the baseline VTimeLLM+CONE model and our \model\ across three Intersection over Union (IoU) thresholds: $\tau_{\text{IoU}=0.1}$, $\tau_{\text{IoU}=0.3}$, and $\tau_{\text{IoU}=0.5}$. Lower ECE values indicate better calibration performance. In this analysis, we set the number of bins, $B=10$. Our model consistently outperforms the baseline across all thresholds, highlighting the effectiveness of calibrated fine-tuning in improving the reliability of confidence estimates. The increase in error with higher IoU thresholds (+0.01\%) is negligible, further validating the robustness of our approach. 

\begin{figure*}
    \centering
    \includegraphics[width=\linewidth]{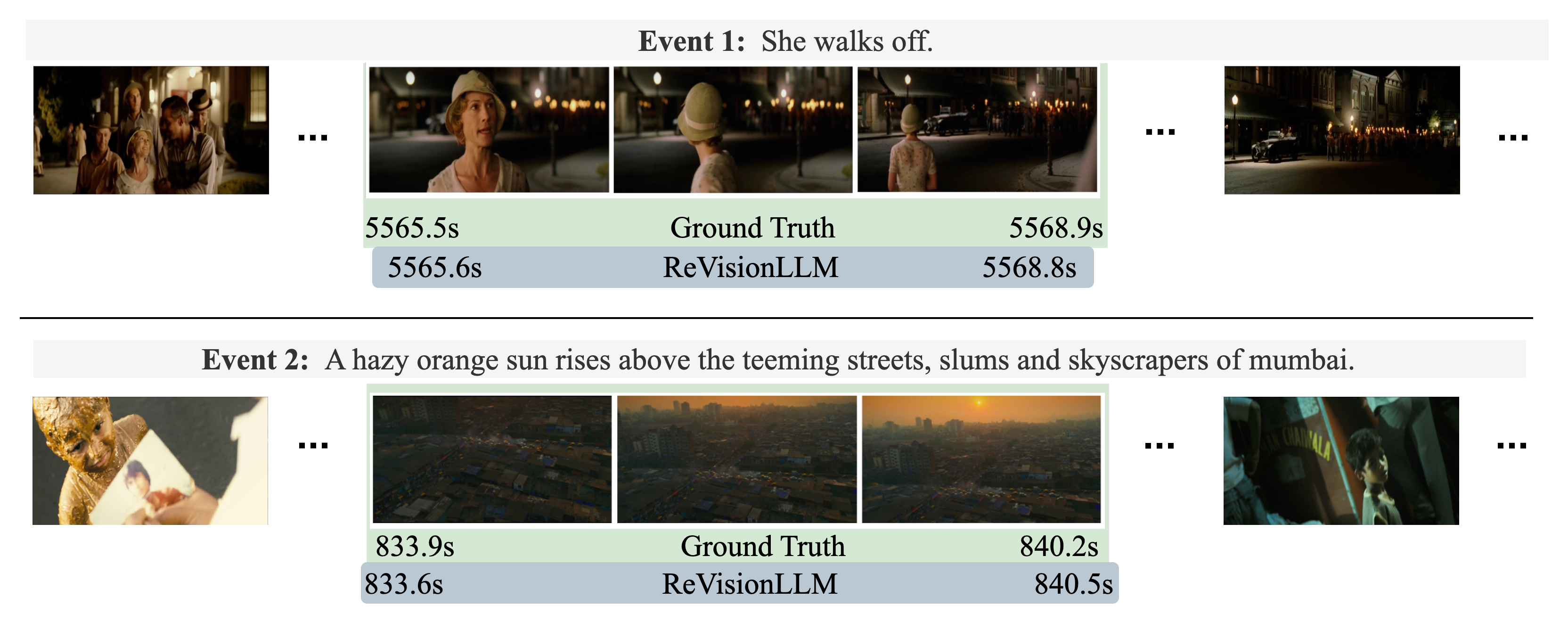}
    \caption{\textbf{Additional Qualitative Results} for Long video temporal grounding on the MAD dataset.  \model\ effectively identifies moments within hour-long movies by leveraging a recursive processing approach that operates at both the short video segment level and hour-long videos. Our VLM baseline completely fails to locate the events in these scenarios.}
    \label{fig:mad}
\end{figure*}
\begin{figure*}
    \centering
    \includegraphics[width=1\linewidth]{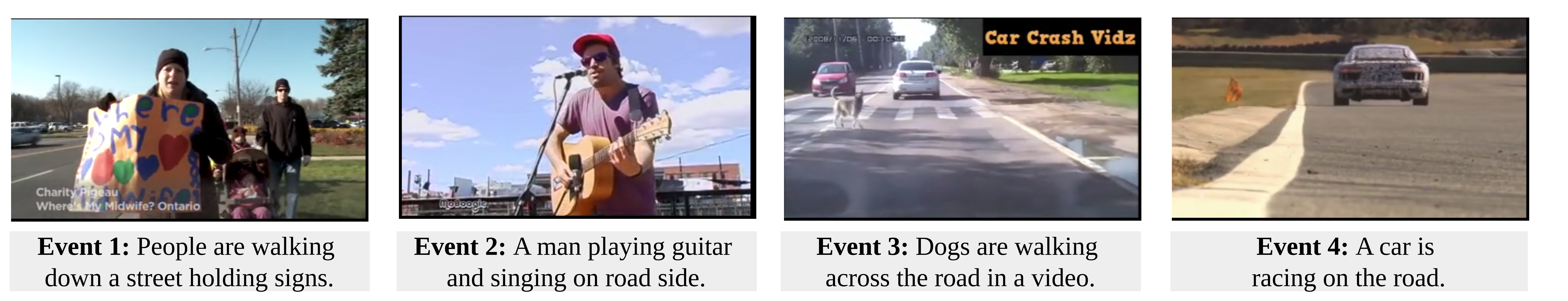}
    \caption{\textbf{Qualitative Results} for Text-to-video retrieval task on MSRVTT dataset. Here, we only show one representative video frame for four diverse queried events, which our model successfully retrieves.}
    \vspace{-2mm}
    \label{fig:msrvtt}
\end{figure*}
\begin{figure*}
    \centering
    \includegraphics[width=\linewidth]{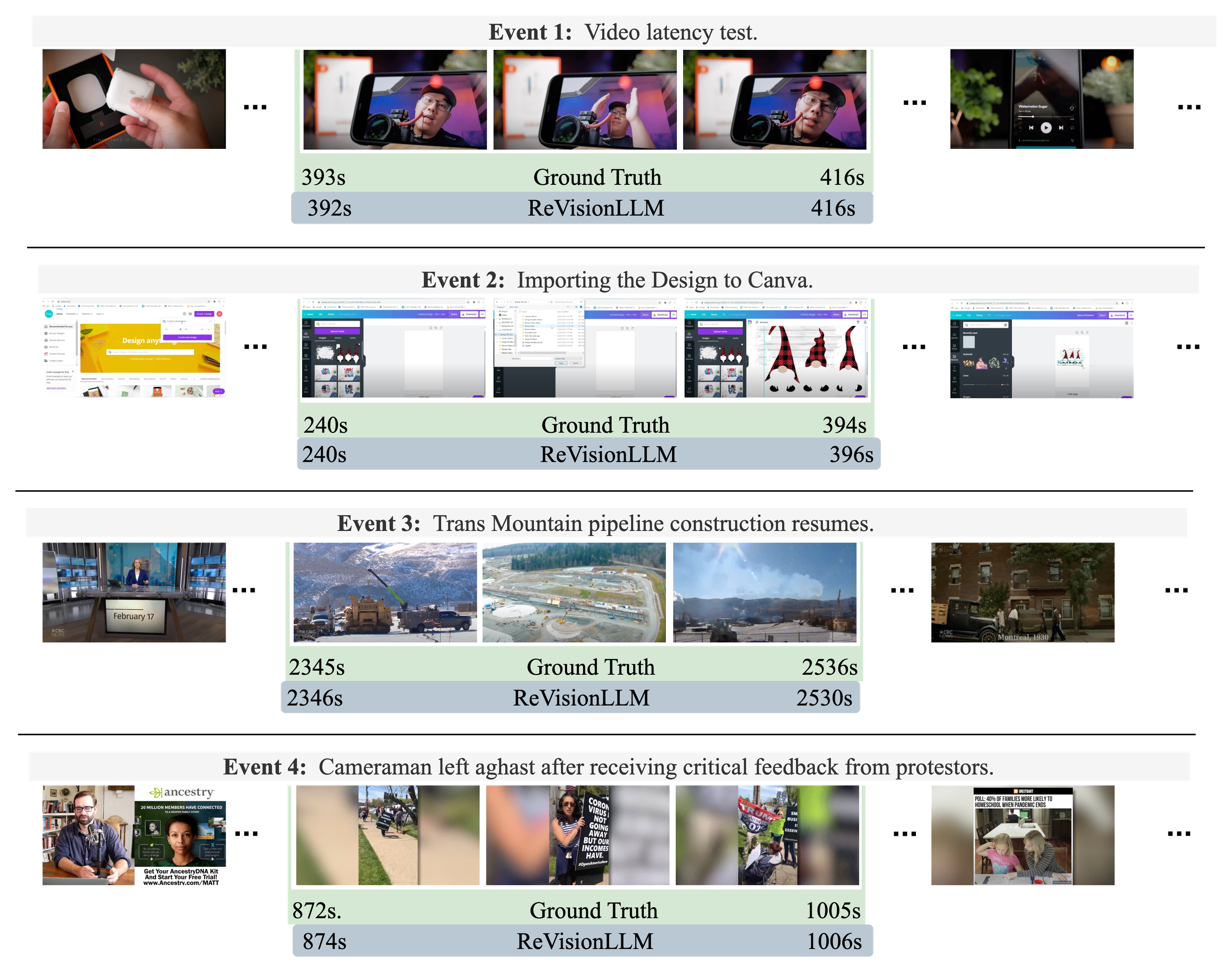}
    \caption{\textbf{Qualitative Results} of long video temporal grounding on the VidChapters-7M dataset. \model\ demonstrates its ability to accurately locate specific events within hour-long YouTube videos across diverse content types, including tutorials, product reviews, news, and podcasts. This precise localization of video chapters has the potential to streamline video search engines and enhance user experience across various online platforms.}
    \label{fig:chapters}
\end{figure*}

\section{Generalization: Text-to-Video Retrieval}
\label{sec:sup generalization}
\noindent\textbf{Problem Statement.}

We show the generalizability of \model\ on the task of text-to-video retrieval, where the goal is to retrieve the most relevant videos from a given video set $\mathcal{V}$ for a provided query text $t$ describing an event. This involves ranking the videos $v \in \mathcal{V}$ based on their similarity to the query. For this problem, the input consists of a video $v$ and a text $t$. We represent a video $v \in \mathbb{R}^{T \times 3 \times H \times W}$ as a sequence of $T$ image frames, where $v = [v^{1}, v^{2}, \ldots, v^F]^T$, and each frame $v^{f}$ has a spatial resolution of $H \times W$ with 3 color channels. Text $S$ is represented the queried sentence with $N_s$ words.

\noindent\textbf{Task Adaptation for \model.} In our original grounding task, we work with a single long video, which we divide into a set of shorter segments, denoted as $C$. In contrast, for the text-to-video retrieval task, we handle multiple videos, forming a set $\mathcal{V}$. To address this, we combine all the videos into one long sequence and predict the index of the relevant video. We uniformly sample 100 frames from each video and use our vision encoder to extract features, resulting in video features $\mathcal{\hat{V}} \in \mathbb{R}^{|\mathcal{V}| \times 100 \times 768}$. These features serve as our video segments, so $C = \mathcal{\hat{V}}$. Additionally, we extract textual features from the query using the text encoder, resulting in $Q \in \mathbb{R}^{N_s \times 768}$, where $N_s$ is the number of words in the query.

For the input prompt, we use: \textit{``\textless video\textgreater~Does the \textless event\textgreater~happen in the video? Answer yes or no."} The model is trained to respond \textit{``Yes."} for relevant videos and \textit{``No."} for irrelevant ones. In this setup, we use the \model-I variant, where we first process each video at the lowest hierarchical level, then revise our predictions recursively at higher levels. At the upper hierarchies, the prompt becomes: \textit{``\textless video\textgreater ~ in which video can we see the \textless event\textgreater ~ happening?"} The model responds with \textit{``In video $v$."}, where $v$ denotes the index of the relevant video.

Finally, we rank the predicted videos based on the calibrated confidence scores from our LLM, ensuring more accurate retrieval results.

\noindent\textbf{Dataset Details.} The MSR-VTT dataset contains 10,000 videos, each associated with around 20 human-annotated captions. Notably, the captions for a single video often describe distinct parts of the content, aligning with our goal of matching a specific textual query to the most relevant frames within a video. The videos in this dataset range in duration from 10 to 32 seconds. For training, we use \textit{9k-Train} split, including approximately 9,000 videos as outlined in \cite{gabeur2020multi}. Unless specified otherwise, our experiments use the \textit{9k-Train} split for training. To evaluate our models, we adopt the \textit{1k-Test} set from \cite{yu2018joint}, which comprises 1,000 carefully selected video caption pairs.

\section{Aditional Qualitative Results} In this section we provide additional qualitative results for MAD, VidChapters-7M and MSRVTT datasets.
\label{sec:sup qualitative} 
\newline \\
\noindent\textbf{MAD Dataset:}
In Figure \ref{fig:mad}, the qualitative results illustrate \model's ability to localize subtle and tiny moments within extremely long videos, often set in visually similar scenes. For Event 1, \model\ accurately identifies the brief instance where a woman walks off amidst a dimly lit street setting, despite the challenge of nearly identical surrounding frames. This demonstrates the model's precision in grounding temporal boundaries in extended sequences where minor actions must be differentiated. In Event 2, the model successfully localizes the moment of a hazy orange sunrise over the sprawling streets, slums, and skyscrapers of Mumbai. This event, embedded within a visually repetitive urban setting, showcases \model's capacity to detect subtle temporal shifts in lighting and atmosphere. These examples emphasize the model's ability to handle the intricacies of long-form videos, identifying precise moments even when scenes exhibit minimal variation, thereby enabling enhanced retrieval and understanding of extended video content. 

\noindent\textbf{MSRVTT Dataset:}
In Figure \ref{fig:msrvtt}, we show a representative video frame for each of the four diverse events that our model successfully retrieved. Events (1) and (2) are visible for short time intervals inside the video, and our model effectively captures these moments due to its ability to focus on fine-grained details. In contrast, events (3) and (4) involve objects with varying speeds and directions, interacting dynamically with their environment. For example, in (3), the dog crosses the road, while in (4), the car moves along the road. These examples demonstrate our model's ability to comprehend and capture both visual and action-related details, enabling it to retrieve the most relevant video from a large dataset.

\noindent\textbf{VidChapers-7M Dataset:}
In Figure \ref{fig:chapters}, the qualitative results from the VidChapters-7M dataset demonstrate \model's ability to enhance online video search and content retrieval across diverse platforms, including YouTube, educational portals, and news archives. In Event 1, \model\ accurately localizes a video latency test in a product review, capturing fine-grained temporal details essential for identifying technical demonstrations. Event 2 showcases the model's ability to navigate complex, sequential workflows, pinpointing design importing actions in Canva. Event 3 highlights \model's proficiency in localizing a news report on Trans Mountain pipeline construction, effectively distinguishing dynamic scenes involving machinery and landscapes—key for indexing news and documentaries. In Event 4, the model identifies a cameraman’s emotional reaction during a podcast, illustrating its capability to understand contextual nuances and interactions. These results emphasize \model's effectiveness in improving content understanding and enabling precise event retrieval in long-form videos, with strong applicability to online video search engines and content recommendation systems.

\end{document}